\documentclass[sigconf]{aamas} 

\usepackage{balance} 

\usepackage{xspace}
\usepackage{algorithm}
\usepackage{algorithmic}
\usepackage{booktabs}
\usepackage{multirow}
\usepackage{subcaption}         
\usepackage{soul}

\usepackage{hyperref}
\usepackage{xr-hyper}

\providecommand{\AAMASARXIVNOTE}{} \providecommand{\LOADAPPENDIXREFS}{}
\LOADAPPENDIXREFS

\doi{XEYY6214}

\makeatletter
\gdef\@copyrightpermission{
  \begin{minipage}{0.2\columnwidth}
   \href{https://creativecommons.org/licenses/by/4.0/}{\includegraphics[width=0.90\textwidth]{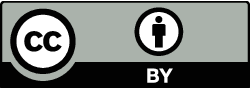}}
  \end{minipage}\hfill
  \begin{minipage}{0.8\columnwidth}
   \href{https://creativecommons.org/licenses/by/4.0/}{This work is licensed under a Creative Commons Attribution International 4.0 License.}
  \end{minipage}
  \vspace{5pt}
}
\makeatother

\setcopyright{ifaamas}
\acmConference[AAMAS '26]{Proc.\@ of the 25th International Conference
on Autonomous Agents and Multiagent Systems (AAMAS 2026)}{May 25 -- 29, 2026}
{Paphos, Cyprus}{C.~Amato, L.~Dennis, V.~Mascardi, J.~Thangarajah (eds.)}
\copyrightyear{2026}
\acmYear{2026}
\acmDOI{}
\acmPrice{}
\acmISBN{}

\acmSubmissionID{25}

\title[AAMAS-2026 Formatting Instructions]{Conservative Equilibrium Discovery in Offline Game-Theoretic Multiagent Reinforcement Learning}

\subtitle{AAAI Track}

\author{Austin A. Nguyen}
\affiliation{
  \institution{University of Michigan}
  \city{Ann Arbor}
  \country{United States}}
\email{ngaustin@umich.edu}

\author{Michael P. Wellman}
\affiliation{
  \institution{University of Michigan}
  \city{Ann Arbor}
  \country{United States}}
\email{wellman@umich.edu}

\begin{abstract}
Offline learning of strategies takes data efficiency to its extreme by restricting algorithms to a fixed dataset of state-action trajectories.
We consider the problem in a mixed-motive multiagent setting, where the goal is to solve a game under the offline learning constraint.
We first frame this problem in terms of selecting among candidate equilibria. 
Since datasets may inform only a small fraction of game dynamics, it is generally infeasible in offline game-solving to even verify a proposed solution is a true equilibrium. 
Therefore, we consider the relative probability of low regret (i.e., closeness to equilibrium) across candidates based on the information available.
Specifically, we extend Policy Space Response Oracles (PSRO), an online game-solving approach, by quantifying game dynamics uncertainty and modifying the RL objective to skew towards solutions more likely to have low regret in the true game.
We further propose a novel meta-strategy solver, tailored for the offline setting, to guide strategy exploration in PSRO.
Our incorporation of \textbf{C}onservatism principles from \textbf{Off}lin\textbf{e} reinforcement learning approaches for strategy \textbf{E}xploration gives our approach its name: \ourapproach.
Experiments demonstrate \ourapproach's ability to extract lower-regret solutions than state-of-the-art offline approaches and reveal relationships between algorithmic components, empirical game fidelity, and overall performance.
\end{abstract}


\newcommand{\BibTeX}{\rm B\kern-.05em{\sc i\kern-.025em b}\kern-.08em\TeX}

\newcommand{\term}[1]{\textbf{\textit{#1}}}
\newcommand{\abs}[1]{\left| #1 \right|}
\newcommand{\E}{\mathbb{E}}
\newcommand{\BR}{\mathit{BR}}
\newcommand{\Regret}{\mathit{Regret}}
\newcommand{\Accept}{\mathit{ACCEPT}}
\newcommand{\rrd}{\mathit{rrd}}
\newcommand{\rrrd}{\mathit{r3d}}
\newcommand{\init}{\mathit{init}}
\newcommand{\eq}{\mathit{eq}}
\newcommand{\eval}{\mathit{eval}}
\newcommand{\train}{\mathit{expl}}
\newcommand{\bc}{\mathit{bc}}
\newcommand{\mix}{\mathit{mix}}
\newcommand{\discount}{\gamma}
\newcommand{\ourapproach}{COffeE-PSRO\xspace}

\newcommand{\uniformDataset}{\pi^{\mathcal{U}}}
\newcommand{\eqMixtureDataset}{\sigma^\eq}

\begin{document}


\pagestyle{fancy}
\fancyhead{}

\maketitle 



\section{Introduction}
Game-theoretic analysis is a powerful tool when evaluating multiagent systems because it focuses attention on configurations where agents behave rationally relative to each other. 
In complex games with large strategy spaces, identifying such equilibrium profiles through \term{multiagent reinforcement learning} (MARL) requires substantial computation and \textit{data}. 
Gathering ample data for game-theoretic analysis can thus be difficult and costly in real-world applications.
A capability to perform such analysis \textit{offline} from pre-generated, fixed-size datasets addresses the extreme case of limits on real-time data collection.

Strategy generation using offline reinforcement learning is well-studied in single-agent and cooperative multiagent domains.
Many offline learning algorithms adopt the \textit{conservatism principle}: strategies should stay in dataset-informed regions and assume unknown regions yield low returns.
This can be interpreted loosely as \textit{policy selection}: among policies that seem approximately optimal, favor ones for which the dataset provides strong evidence and thus better relative confidence of high actual (true) return.
Analogously, we frame offline game-solving as \textit{equilibrium selection}, where we choose among profiles apparently in approximate equilibrium, aiming to minimize actual (true-game) regret.

Our contribution is in the application and extension of conservatism to offline \textit{game-solving}.
This extension poses a special challenge: consideration of a candidate strategy profile's potential strategic deviations.
Our approach achieves this by combining offline RL algorithmic concepts and Policy Space Response Oracles (PSRO) \citep{lanctot2017unified_psro}, an online game-solving framework that iteratively extends a strategy set using deep RL best-responses to mixed strategies in an empirically estimated game model.
Given a dataset of trajectories (sequences of true-game multiagent interactions), we first train an ensemble dynamics model to replace an online simulator and use prediction differences within the ensemble to quantify uncertainty under the dataset.
We apply conservatism by introducing a response objective that balances reward, uncertainty given the current response target, and uncertainty of potential strategic deviations. 
Further, we propose a novel meta-strategy solver that minimizes a heuristic-based, pessimistic estimate of regret.
This extension of \textbf{C}onservatism from \textbf{Off}lin\textbf{e} RL approaches to aid strategy \textbf{E}xploration in PSRO gives our approach its name: \ourapproach.
In experiments on a sequential bargaining game, we find that \ourapproach tends to output lower-regret solutions than tested baselines.
Our results highlight the role of conservatism in improving model fidelity, robustness to dataset changes, and solution regret.

\section{Related Work}

Conservatism is a common theme in offline RL approaches \citep{levine2020offline_tutorial}.
Model-free methods have incorporated conservatism through policy regularization \citep{kumar2019_bear, wu2019_brac}, value penalties \citep{kumar2020_cql}, and implicit behavioral constraints \citep{nair2020awac, peng2019advantage_awr}, while model-based methods have used uncertainty quantification and reward penalties \citep{yu2020_mopo, kidambi2020_morel}.
Conservative methods in offline MARL have addressed centralized coordination \citep{wang2024offline_global_implicit} and nonstationarity in decentralized optimization \citep{jiang2023offline_transition_normalization}.

\citet{yang2020overview_game_theoretic_marl} provide an overview of game-theoretic approaches in MARL.
The framework of \term{Policy Space Response Oracles} (PSRO) \citep{Bighashdel24,lanctot2017unified_psro} employs an empirical game modeling approach \citep{Wellman25tg}, iteratively extending the strategy set using deep RL to derive policies maximizing return against an other-player \term{response target} defined by a \term{meta-strategy solver} (MSS).
Substantial work has investigated strategy exploration \citep{Jordan10sw} in PSRO, in particular through MSS selection of response targets \citep{balduzzi2019open_rectified_nash, marris2021multi_joint_psro, muller2020generalized_alpha_psro, smith2021iterative_qmix, wang2023regularization_rrd} or through definition of response objectives \citep{Perez-Nieves21,wang2024generalized_gro, ex2psronguyenexplicit}. 

In offline game-theoretic MARL, theoretical work has investigated the necessary conditions for equilibrium extraction \citep{cui2022offline_solvable, zhong2022pessimistic}, defining \textit{unilateral concentration} as the property where a dataset covers an equilibrium and all its unilateral strategic deviations. 
Other approaches focus on offline evaluation of given strategies to determine which are most likely equilibria \citep{zhang2023_offline_markov_games_general_function, zhong2022pessimistic}.
\citet{shao2024copsro} employ conservative critics and low-rank matrix completions to select equilibria given a fixed dataset of payoff data for select policy profiles.
Most relevant here is work on strategy generation for offline equilibrium extraction \citep{li2022offline_equilibrium_finding}, which uses a trained dynamics model in place of a simulator and mixes the final solution with a behavior cloning policy.
\citet{chen2024offline_fictitious_self_play} adapts fictitious self-play using single-agent offline RL and importance sampling.

\section{Notation}

We model a multiagent environment as a \term{stochastic game} defined by $\langle \mathcal{S}, \mathcal{A}, \mathcal{T}, \mathcal{R}, Z, O, n, \gamma \rangle$.
The environment follows trajectories of states $s \in \mathcal{S}$, influenced by actions $a \in \mathcal{A}$ taken alternately by each of $n$ agents. 
Each action prompts a transition according to the state transition function $\mathcal{T}(s' \mid s, a): \mathcal{S} \times \mathcal{A} \times \mathcal{S} \rightarrow [0, 1]$. 
Agent rewards are given by $\mathcal{R}(s, a): \mathcal{S} \times \mathcal{A} \rightarrow \mathbb{R}^n$, and $\gamma \in [0, 1)$ is a discount factor.
The environment is partially observable, meaning agents draw observations $z \in Z$ from an observation function $O(s): \mathcal{S} \rightarrow Z$. 
We denote an action-observation history by~$h_t$, with $\mathcal{H}_t \equiv (Z \times \mathcal{A})^t$ the space of conceivable $t$-length histories.
A \term{policy} or \term{pure strategy} $\pi_i$ for player~$i$ is defined as a mapping from $h_t$ to a probability distribution over legal actions $\pi_i: \mathcal{H} \to \Delta(\mathcal{A})$.
We omit subscript $i$ to refer to a \term{joint policy} (or \term{strategy profile}) $\pi=(\pi_1,\dotsc,\pi_n)$.
A \term{trajectory} $\tau$ describes a rollout of game play and is represented by a sequence of tuples $\tau = \left((s_t, o_{i,t}, a_{i,t}, 
r_{i,t}, i)\right) _{t=1}^{\abs{\tau}}$, where~$t$ indexes the timestep.
A dataset $\mathcal{D}$ is a collection of trajectories $\{\tau^j\}_{j=1}^{\abs{\mathcal{D} }}$, indexed here by~$j$.
We may subscript a trajectory $\tau_\pi^j$ to indicate the joint policy $\pi$ generating the rollout.
The joint probability of being in state~$s$ and executing action~$a$ under joint policy~$\pi$ is denoted by $\mu^{\pi}(s,a)$.

Multiagent interactions may also be expressed in \term{normal form}.
A normal form \term{symmetric game} $\Gamma$ is represented by a tuple $(\Pi,U,n)$, where $n$ is the number of players, $\Pi = \Pi_1\times\dotsm\times\Pi_n$ is a set of joint strategies ($\pi \in \Pi$), and \term{payoff function} $U:\Pi\to\mathbb{R}^n$ maps strategy profiles $\pi$ to vectors of payoffs (i.e., returns).
A probability distribution $\sigma_i \in \Delta(\Pi_i)$ over pure strategies is called a \term{mixed strategy}.
We also use~$\sigma_{i, s}$ to indicate player~$i$'s strategy applied at step $s$.
A subscript $-i$ indicates all players except~$i$.
We write player $i$'s payoff for playing $\pi_i$ when the others play $\pi_{-i}$ as $u_i(\pi_i,\pi_{-i})$, where $u$ is a length-$n$ utility vector.
For mixed strategies, the payoffs are given by expectation:
\begin{align*}\label{eq:exp-payoff}
    u_i(\pi_i,\sigma_{-i}) &= \E_{\pi_{-i}\sim \sigma_{-i}}u_i(\pi_i, \pi_{-i}),\\
   u_i(\sigma) &= \E_{\pi_i\sim \sigma_i}u_i(\pi_i, \sigma_{-i}).
\end{align*}
Player $i$'s \term{best response} to all others playing $\sigma$ is given by $\BR_i(\sigma) = \arg\max_{\sigma_i^*\in\Delta(\Pi_i)} u_i(\sigma_i^*,\sigma_{-i})$.
Mixed strategy $\sigma^*$ is a \term{Nash equilibrium} (NE) iff $\forall i.\ \sigma_i^* \in \BR_i(\sigma^*)$.
The potential gain to deviating from~$\sigma$ is called \term{regret}: $\Regret_i(\sigma)= u_i(\pi_i^*,\sigma_{-i}) - u_i(\sigma)$, where $\pi_i^*\in \BR_i(\sigma)$. 
By definition, if $\sigma^*$ is a NE, its regret is zero. 
An \term{empirical game model} $\hat{\Gamma}=(\hat{\Pi}, \hat{U}, n)$ is a normal-form game over a finite joint-strategy subspace $\hat{\Pi} = \hat{\Pi}_1\times\dotsm\times\hat{\Pi}_n$, $\hat{\Pi}_i\subset \Pi_i$, with payoff function~$\hat{U}$ estimated through simulation.

\section{Conservatism in Offline Game-Solving}

Empirical game updates and best-response training in PSRO require explicit access to a game simulator.
The offline setting, in contrast, works with fixed datasets of unspecified quality. 
Assessing candidate solutions requires reliable information on the included strategies \textit{and} relevant deviations, which is contingent on the dataset's behavior coverage, or lack thereof.
Since we cannot draw assumptions on the extent of this coverage, offline game-solving must account for uncertainty in regret assessments.
A joint strategy $\pi$ is considered \term{covered}
if the dataset contains all state-action pairs visitable under $\pi$: $\forall s,a.\ \mu^{\pi}(s, a) > 0 \implies (s, a) \in \mathcal{D}$.
In large games, full coverage of any policy $\pi$ and its unilateral deviations is infeasible.
Therefore, we consider the \textit{extent} to which a policy $\pi$ is covered.
Our approach frames offline game-solving as an equilibrium selection problem: among solutions that seem to have low regret offline, we prefer those more certain to have low regret in the true game. 
By skewing offline strategy exploration toward high-confidence policy spaces, we seek to produce lower-regret solutions than alternative exploration methods with some consistency.


\subsection{Training the Ensemble Dynamics Model}
\label{sec:train_dynamics_model}

Our model-based approach adapts analogous single-agent offline RL approaches. 
We train an ensemble of $K$ multi-layer perceptrons as  dynamics models, each estimating the transition function~$\mathcal{T}$, reward function $\mathcal{R}$, observation mapping $\mathcal{O}$, and state-dependent action space $\mathcal{A}$, 
from which dataset $\mathcal{D}$ was generated.
Henceforth, our references to these MDP components denote predictions of a learned dynamics model. 
Ensemble predictions are the average prediction across all $K$ networks: $\overline{\mathcal{R}}_i(s_t, a_t) = \frac{1}{K}\sum_{j=1}^K \mathcal{R}_i^j(s_t, a_t)$, where $\mathcal{R}^j_i(s_t, a_t)$ refers to model $j$'s prediction of player $i$'s reward given state-action pair ($s_t, a_t$), and so forth for other MDP components. 
Additional details, including terminal handling, information state generation, and training parameters are provided in Supp.~\ref{sec:train_dynamics_model} \AAMASARXIVNOTE.

Prediction differences between intra-ensemble models can serve as proxy for uncertainty quantification \citep{kidambi2020_morel}.
We follow \citet{yu2020_mopo} and define $\rho$ based on the range of model predictions:%
\footnote{The original approach of \citet{yu2020_mopo} used transition function differences.
We instead employ reward differences, which achieved stronger results in our experiments.}
\begin{equation}
    \rho(s_t, a_t, \mathcal{R}) = \max_{j, k} \sum_{i=1}^n\abs{ \mathcal{R}_i^j(s_t, a_t) - \mathcal{R}_i^k(s_t, a_t) }.
    \label{eq:discrepancy_definition}
\end{equation}
Equation~\ref{eq:discrepancy_definition} reflects an assumption that models tend to provide different reward predictions on information-deprived transitions, and similar predictions for sufficiently informed transitions (whether informed explicitly or implicitly by dataset $\mathcal{D}$).
The maximum prediction difference is a conservative indicator \citep{yu2020_mopo, kidambi2020_morel}.

\subsection{A Conservative Response Objective}

\begin{figure*}[hbt]
    \centering
    \includegraphics[width=1.0\linewidth]{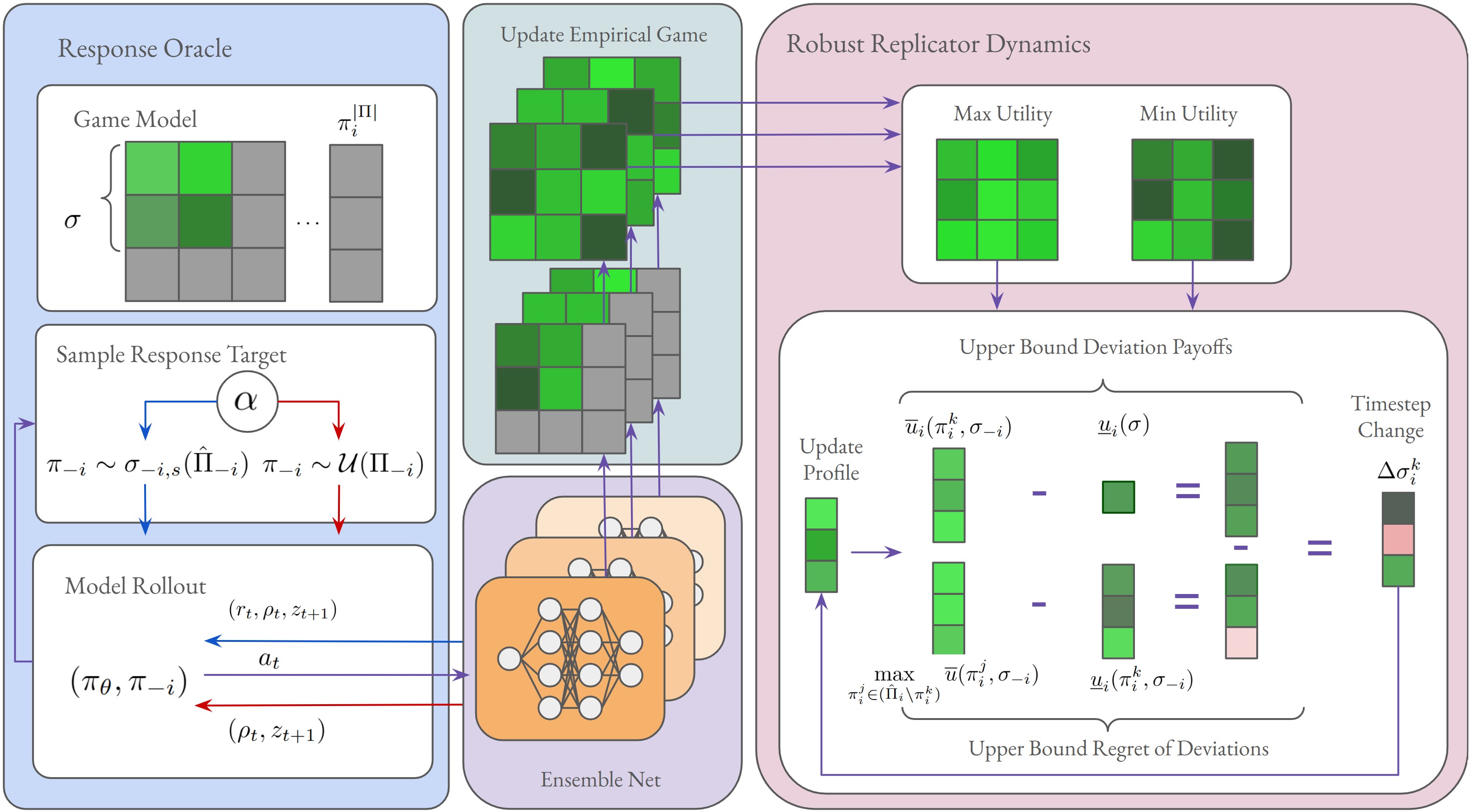} 

    \caption{An iteration of \ourapproach. 
    During strategy generation, a rollout optimizes $\rho$ against a uniformly sampled policy with probability $\alpha$, or the conservative response to the target (Equation~\ref{eq:penalty_response_objective}) otherwise.
    The two cases are indicated by red and blue arrows.
    The newly trained policy is added to the strategy set, updating the empirical game using pessimistic and optimistic utility estimates. 
    R2D determines the next iteration's response target.}
    \label{fig:overall_algorithm_schematic}
    \Description{Coffee-PSRO algorithm schematic.}
\end{figure*}

At each iteration of online PSRO, strategies are trained by optimizing a single agent's expected return against a fixed, stationary response target, $\sigma_{-i}$. 
\begin{align}
   \pi_i &= \arg\max_{\pi_{\theta}} J_i(\theta, \sigma_{-i}),\nonumber\\
    J_i(\theta, \sigma_{-i}) &= \mathbb{E}_{a_t \sim (\pi_{\theta}, \sigma_{-i})}\sum_{t=1}^T  \overline{\mathcal{R}}_i(s_t, a_t).\label{eq:psro_response_objective}
 \end{align}
Under a fixed dataset, we cannot assume availability of covered best-responses to $\sigma_{-i}$.
\ourapproach modifies PSRO's response objective by optimizing an augmented reward function, $J_i'(\theta, \sigma_{-i})$
\begin{equation}
     = \mathbb{E}_{a_t \sim (\pi_{\theta}, \sigma_{-i})}\left[\sum_{t=1}^T  \overline{\mathcal{R}}_i(s_t, a_t) - \lambda \rho(s_t, a_t, \mathcal{R})\right],\label{eq:penalty_response_objective}
 \end{equation}
where $\lambda$ is a hyperparameter. 
Equation~\ref{eq:penalty_response_objective} incorporates model-based uncertainty given $\sigma_{-i}$.
Although this objective is sufficient in single-agent RL problems, offline game-solving requires information on strategy deviations to additionally \textit{verify} equilibrium properties \citep{cui2022offline_solvable}.
We hypothesize that skewing strategy exploration toward equilibria with low $\rho$ on \textit{unilateral deviations} will also improve performance.
Suppose player $i$ is creating a strategy exclusively optimizing for unilateral concentration.
This entails creating a strategy $\pi_i$ that provides other players the least uncertainty $\rho$ on all potential responses to~$\pi_i$.
Fig.~\ref{fig:overall_algorithm_schematic} demonstrates we can achieve this by sampling a different other-player policy at each training episode and minimizing $\rho$. 
A roughly equivalent formulation%
\footnote{Equivalence requires that the uniform distribution over strategies be behaviorally equivalent to the uniform-random strategy, which does not generally hold. 
In $N$-player games, only one player can use uniform-random action at any given episode.} is to respond to a uniform-random action policy exclusively optimizing $\rho$:
\begin{multline}
        J_i''(\theta, \sigma_{-i}) = (1 - \alpha) J_i'(\theta, \sigma_{-i}) \\
        -\alpha \mathbb{E}_{a_t\sim(\pi_\theta, \mathcal{U}(\mathcal{A}))} \sum_{t=1}^T \lambda\rho(s_t, a_t, \mathcal{R}),
    \label{eq:overall_response_objective}
\end{multline}
where $\alpha$ is a hyperparameter denoting the probability of optimizing strategic deviation coverage in each training episode. 
In practice, we linearly anneal $\alpha$ to zero over 
PSRO iterations.
In the context of strategy generation, this objective balances model-predicted reward, uncertainty given the response target $\sigma_{-i}$, and uncertainty of potential other-player deviations.
Under the lens of equilibrium selection, Equation~\ref{eq:overall_response_objective} can be viewed as skewing strategy exploration toward higher-certainty policy regions \citep{wang2024generalized_gro}.

\subsection{A Conservative Meta-Strategy Solver}

We propose a novel MSS that uses ensemble model predictions to derive a conservative response target.
This MSS takes concepts from the \term{Bellman-consistent equilibrium learning} (BCEL) method of \citet{zhang2023_offline_markov_games_general_function}, incorporated in a game-solver based on \term{replicator dynamics} (RD) \citep{RD}.
BCEL employs lower and upper utility bounds for given joint strategies and identifies the solution minimizing worst-case regret. 
We design an RD update that computes a stable point in heuristically estimated, worst-case regret.
We define the utility predicted by model~$j$ for player~$i$ under joint policy~$\pi$ as $u_i^j(\pi) = \frac{1}{N} \sum_{k=1}^N \sum_{(s_t, a_t) \in \tau^k_{\pi}} \mathcal{R}_i^j(s_t, a_t)$, where returns are averaged over $N$ trajectories $\tau^k_\pi$. 
Lower and upper utility bounds are replaced by the minimum and maximum utility estimates over the ensemble $\mathcal{R}$, respectively: $ \overline{u}_i(\pi) = \max_{j \in \{1 \dotsc K\}} u_i^j(\pi)$.
Pessimistic utilities $\underline{u}(\pi)$ are minimized over $K$ models. 
We modify the RD update by myopically optimizing for the current pessimistic regret estimate at each step:
\begin{align*}
    \mathit{UBDP} &\doteq \overline{u}_i(\pi_i^k, \sigma_{-i})   - \underline{u}_i(\sigma)\\ 
    \mathit{UBDR} &\doteq \max_{\pi^j_i \in (\hat{\Pi}_i \backslash \pi_i^k)}\overline{u}(\pi^j_i, \sigma_{-i}) - \underline{u}_i(\pi_i^k, \sigma_{-i}) \\
    \Delta\sigma_i^k &= \sigma_i^k [\mathit{UBDP} - \mathit{UBDR}]
    \label{eq:r2d_update}
\end{align*}
%
%
%
where $\sigma^k_i$ is the probability player $i$ plays strategy $k$ and $\Delta$ indicates the amount of change applied to a variable. 
The first term captures an upper-bound deviation payoff (UBDP) for playing $\pi^k_i$ against $\sigma_{-i}$. 
The second represents an upper-bound deviating regret (UBDR) if the player deviates to $\pi^k_i$. 
Our approach accounts for robust offline evaluation, giving our MSS its name: \term{Robust Replicator Dynamics} (R2D).

Neither RD nor R2D offer theoretical convergence guarantees.
Nevertheless, in practice, RD reliably reaches exact or approximate equilibria in empirical games \citep{Wellman25tg} with demonstrated success within the PSRO framework.
Analogously, we demonstrate R2D's empirical benefits within \ourapproach.
A complementary RD variant, Regularized RD (RRD) \citep{wang2023regularization_rrd}, employs an early-stop condition based on a regret threshold to avoid overfitting to the current empirical game solution.
R2D can analogously adopt a stopping threshold measured against worst-case regret, resulting in another MSS called \term{Robust Regularized Replicator Dynamics} (R3D).

\begin{algorithm}[!htb]
\caption{Conservative Offline Exploration PSRO (\ourapproach)}
\label{alg:overall_algorithm}
\raggedright
\textbf{Input}: Dataset $\mathcal{D}$, Penalty weight $\lambda$, Coverage weight $\alpha_{\init}$, Steps anneal coverage weight $S_{\alpha}$, meta-strategy solver $\mathcal{M}$, PSRO iterations $S$, Number of simulations $N$ \\
\textbf{Output}:  Player strategy sets $\hat{\Pi}$, Player profiles $\sigma_S$
\begin{algorithmic}[1] 
\STATE Initialize empty empirical game $\hat{\Gamma}: (\hat{\Pi}, \hat{U}, n)$
\STATE Initialize strategy sets $\hat{\Pi}_i\gets\{ \pi^0_i\}, i\in\{1\dotsc n\}$ 
\STATE \textit{Train ensemble dynamics model of size $K$ on $\mathcal{D}$}
\STATE Update $\hat{\Gamma}$ \textit{using dynamics model} with $N$ simulations
\FOR{each PSRO iteration $s$ from 1 to $S$}
    \FOR{each player $i$ from 1 to $n$}
        \STATE $\pi_i \leftarrow \arg\max_{\pi_{\theta}} J_i''(\theta, \sigma_{-i})$  \eqref{eq:overall_response_objective}, using $\lambda$ and $\alpha$
        \STATE Update strategy sets $\hat{\Pi}_i\gets\hat{\Pi}_i \cup \{\pi_i\}$
    \ENDFOR
    \STATE Update $\hat{\Gamma}$ \textit{using dynamics model} with $N$ simulations
    \STATE Update joint profile $\sigma_s\gets\mathcal{M}(\hat{\Gamma})$
    \STATE \textit{Anneal $\alpha\gets\max(0, \frac{\alpha_{\init}(S_{\alpha} - i)}{S_{\alpha}})$}
\ENDFOR
\STATE \textbf{return} $\hat{\Pi}$, $\sigma_S$
\end{algorithmic}
\end{algorithm}

\subsection{Conservative Offline Exploration in PSRO (COffeE-PSRO)}

Algorithm~\ref{alg:overall_algorithm} summarizes \ourapproach, depicted schematically in Fig.~\ref{fig:overall_algorithm_schematic}.
\ourapproach follows the basic structure of PSRO, using a trained dynamics model in place of the online simulator, and using Equation~\ref{eq:overall_response_objective} for strategy generation.
Any meta-strategy solver $\mathcal{M}$ can be used with \ourapproach,%
\footnote{For conventional MSSs, empirical game payoffs would be estimated using average reward predictions rather than bounds.}
but our experiments suggest that R2D yields the best results.

\section{Experiments}

\subsection{Bargaining}

We test our approach by adapting a 2-player, turn-based, imperfect information, symmetric bargaining game \citep{bargaining}. 
The game (which we refer to as \textit{Bargaining}) features a pool of items that players must divide between them. 
Each player has a randomly drawn, private valuation that gives their per-unit utility value for each item. 
Players alternate making offers of division until one accepts or the game reaches a maximum number of iterations. 
If a player accepts, players receive utility according to the accepted allocations and respective value functions, discounted by $\discount^t$ where $0 < \discount < 1$ is a hyperparameter and $t$ is the number of bargaining rounds so far.
We set $\discount$ to $.99$ 
If no agreement is reached by the turn limit ($T=10$), both players receive zero utility.
Supp.~\ref{sec:appendix_bargaining} provides further details.

\begin{table*}[h]
    \centering

    \caption{Regret of final solutions, averaged over 10 trials with corresponding standard deviation and P-values, each consisting of $S=40$ iterations.
    Bolded algorithmic components indicate our contributions. Datasets contain a varied number of trajectories $N \in \{500, 1000, 2000\}$ and are generated by two different behavior policies: uniform random $\pi^{\mathcal{U}}$, and a mixture over online-generated equilibria $\sigma^{\eq}$. We distinguish between the MSS $\mathcal{M}_{\train}$ used during training versus $\mathcal{M}_{\eval}$ during evaluation to isolate effects on strategy exploration and immediate regret \citep{wang2022evaluating}. All P-values are calculated with Welch's t-test to determine whether \ourapproach+R2D yields significant regret differences relative to the corresponding approach. True-game regret evaluations are discussed in Supp.~\ref{sec:true_game_regret_calculations}.}

         
    
    \begin{tabular}{l@{ \hspace{.25cm}}c@{\hspace{.2cm}}c@{\hspace{.2cm}}c@{\hspace{.2cm}}c@{\hspace{.2cm}}c@{\hspace{.2cm}}c@{}}
        \toprule
        Approach & (500, $\eqMixtureDataset$) & (500, $\uniformDataset$) & (1000, $\eqMixtureDataset$) & (1000, $\uniformDataset$) & (2000, $\eqMixtureDataset$) & (2000, $\uniformDataset$) \\
        \midrule 
        Alg, $\mathcal{M}_{\train}$ &\multicolumn{6}{c}{$\mathcal{M}_{\eval}$: RD} \\
        \midrule
        \textbf{\ourapproach, R2D} & 
            $\textbf{2.02} \pm \textbf{0.40} \mid $ N/A& 
            $4.23 \pm 0.77 \mid $ N/A& 
            $\textbf{1.10} \pm \textbf{0.32} \mid $ N/A & 
            $\textbf{2.40} \pm \textbf{0.55} \mid $ N/A& 
            $\textbf{0.88} \pm \textbf{0.35} \mid $ N/A& 
            $\textbf{1.66} \pm \textbf{0.57} \mid $ N/A\\
        \textbf{\ourapproach}, RD & 
            $2.58 \pm 0.40 \mid$ .01& 
            $4.20 \pm 0.81 \mid$ .53 & 
            $1.47 \pm 0.74 \mid$ .09& 
            $2.64 \pm 0.55 \mid$ .17 & 
            $1.64 \pm 0.94 \mid$ .02 & 
            $3.15 \pm 2.09 \mid$ .03 \\
        OEF, \textbf{R2D} & 
            $2.71 \pm 0.85 \mid$ .02 & 
            $5.01 \pm 1.07 \mid$ .04 & 
            $1.42 \pm 0.52 \mid$ .05 & 
            $3.20 \pm 0.37 \mid$ .01 & 
            $1.03 \pm 0.67 \mid$ .27 & 
            $2.24 \pm 0.84 \mid$ .01 \\
        OEF, RD & 
            $2.38 \pm 0.55 \mid$ .06 & 
            $5.09 \pm 1.19 \mid$ .04 & 
            $1.19 \pm 0.29 \mid$ .29 & 
            $3.74 \pm 2.10 \mid$ .04 & 
            $1.40 \pm 0.79 \mid$ .04 & 
            $3.28 \pm 2.16 \mid$ .02 \\
        OEF-BC, \textbf{R2D} & 
            $2.91 \pm 0.61 \mid$ .01 & 
            $4.00 \pm 0.65 \mid$ .76 & 
            $2.00 \pm 0.22 \mid$ .01 & 
            $2.62 \pm 0.22 \mid$ .13 & 
            $1.85 \pm 0.29 \mid$ .01 & 
            $2.15 \pm 0.23 \mid$ .01 \\
        OEF-BC, RD & 
            $2.82 \pm 0.38 \mid$ .01 & 
            $\textbf{3.85} \pm \textbf{0.54} \mid$ .89 & 
            $1.92 \pm 0.19 \mid$ .01 & 
            $2.96 \pm 1.07 \mid$ .08 & 
            $1.92 \pm 0.49 \mid$ .01 & 
            $2.83 \pm 1.12 \mid$ .01 \\
        \midrule
        Alg, $\mathcal{M}_{\train}$ &\multicolumn{6}{c}{$\mathcal{M}_{\eval}$: \textbf{R2D}} \\
        \midrule 
       \textbf{\ourapproach, R2D} & 
           $\textbf{1.96} \pm \textbf{0.45} \mid $ N/A & 
           $3.98 \pm 0.60 \mid $ N/A& 
           $1.22 \pm 0.41 \mid $ N/A& 
           $\textbf{2.12} \pm \textbf{0.27} \mid $ N/A& 
           $\textbf{0.64} \pm \textbf{0.12} \mid $ N/A& 
           $\textbf{1.40} \pm \textbf{0.36} \mid $ N/A\\
        \textbf{\ourapproach}, RD & 
            $2.15 \pm 0.43 \mid$ .17 & 
            $\textbf{3.76} \pm \textbf{0.75} \mid$ .76 & 
            $1.34 \pm 0.42 \mid$ .26 & 
            $2.85 \pm 0.91 \mid$ .02 & 
            $1.30 \pm 0.59 \mid$ .01 & 
            $2.42 \pm 1.43 \mid$ .05 \\
        OEF, \textbf{R2D} & 
            $2.47 \pm 0.82 \mid$ .05 & 
            $4.90 \pm 1.11 \mid$ .02 & 
            $1.08 \pm 0.37 \mid$ .78 & 
            $3.25 \pm 1.00 \mid$ .35 & 
            $0.73 \pm 0.52 \mid$ .28 & 
            $1.93 \pm 0.49 \mid$ .01 \\
        OEF, RD & 
            $2.11 \pm 0.39 \mid$ .22 & 
            $4.88 \pm 1.03 \mid$ .02 & 
            $\textbf{1.01} \pm \textbf{0.19} \mid$ .92 & 
            $3.68 \pm 1.63 \mid$ .01 & 
            $0.96 \pm 0.59 \mid$ .06 & 
            $2.79 \pm 1.98 \mid$ .03 \\
        OEF-BC, \textbf{R2D} & 
            $2.77 \pm 0.33 \mid$ .01 & 
            $3.97 \pm 0.65 \mid$ .51 & 
            $1.89 \pm 0.16 \mid$ .01 & 
            $2.70 \pm 0.39 \mid$ .01 & 
            $1.68 \pm 0.22 \mid$ .01 & 
            $2.79 \pm 1.98 \mid$ .03 \\
        OEF-BC, RD & 
            $2.65 \pm 0.20 \mid$ .01 & 
            $3.79 \pm 0.46 \mid$ .78 & 
            $1.98 \pm 0.19 \mid$ .01 & 
            $2.92 \pm 0.75 \mid$ .01 & 
            $1.82 \pm 0.18 \mid$ .01 & 
            $2.65 \pm 0.94 \mid$ .01 \\
        \bottomrule
    \end{tabular}

    \label{fig:main_regret_plots}
\end{table*}

\subsection{Implementation and Reproducibility}

\label{sec:implementation_and_reproducibility}

Our experiments build on an existing codebase, Deepmind's OpenSpiel \citep{openSpiel}.
We refer to datasets using tuples $(N, \pi^\beta)$ where $N$ is the number of trajectories and $\pi^\beta$ is the joint behavior policy.
Our experiments vary $N \in \{500, 1000, 2000\}$ and $\pi^{\beta} \in \{\uniformDataset, \eqMixtureDataset\}$, where $\uniformDataset$ is a uniform-random action policy, and $\eqMixtureDataset$ is a uniform mixture over 5 equilibrium policies generated by online PSRO (Supp.~\ref{sec:dataset_generation}).
Under these settings, coverage of Bargaining's full strategy space is highly unlikely, providing benchmarks that measure ability to extract approximate equilibria under limited information.
All trials use Double Deep Q-Networks (DDQN) \citep{van2016deep} for best-response calculations, yielding deterministic, pure strategies $\pi_i$. 
Supp.~\ref{sec:hyperparameters} discusses hyperparameters for our best-response and dynamics model.

We compare our approach to two benchmarks, both variants of Offline Equilibrium Finding PSRO \citep{li2022offline_equilibrium_finding}.
These approaches train a dynamics model, querying it in place of an online simulator (equivalent to \ourapproach with $\lambda = \alpha = 0$), and either use the output solution, or mix that solution with a behavior-cloned policy $\pi^{\bc}_i$ on $\mathcal{D}$ by weight $\alpha_{\bc}$. 
We refer to the former as OEF and the latter as OEF-BC.
All approaches use symmetric implementations, where players share a strategy set~$\hat{\Pi}_i$ and profile~$\sigma_i$, $\hat{\Gamma}$ is a symmetric game, and a single policy~$\pi_{i}$ is generated at each iteration. 
Analogous to how offline RL uses online simulations for return evaluation, we use online best responses (detailed in Supp.~\ref{sec:true_game_regret_calculations}) to evaluate true-game regret.
All trials consist of $S=40$ iterations, each iteration corresponding to one strategy generation.
We tune \ourapproach and OEF-BC using $\mathcal{D} = (1000, \uniformDataset)$.%
\footnote{OEF has no additional parameters.}
\ourapproach is tuned across ($\lambda$, $\alpha$) for both $\mathcal{M} \in \{\text{RD}, \text{R2D}\}$, running 5 trials for each, and an additional 5 for the initially lowest regret configuration. 
OEF-BC is tuned exhaustively across $\alpha_{\bc}$, where all algorithmic variants immediately run 10 trials and then select the lowest regret parameter setting.
We fix tuned parameters across datasets.
Note that this tuning scheme is more exhaustive than \ourapproach's.
We chose to use R2D as opposed to R3D in order to isolate the effects of optimizing for conservative regret estimates over expected regret.%
\footnote{This additionally removes the need to tune additional parameters $\lambda_{\rrd}$ and $\lambda_{\rrrd}$ for fair comparison.}
Hyperparameters are detailed in Supp.~\ref{sec:hyperparameters}.

\subsection{Regret Analysis}

Table~\ref{fig:main_regret_plots} compares tuned versions of \ourapproach, OEF, and OEF-BC. 
We isolate and measure the improvement from an MSS used in strategy exploration $\mathcal{M}_{\train}$ by using a consistent \textit{evaluation} MSS $\mathcal{M}_{\eval}$ across different approaches \citep{wang2022evaluating}.
Results suggest that separately applying \ourapproach or R2D yields variable improvement. 
However, the combined use of \ourapproach with R2D tends to outperform tested baselines, suggesting a strong, beneficial interplay between conservatism in strategy generation and MSS.

\begin{figure*}[htb]
    \centering
    \begin{subfigure}[t]{.32\linewidth}
        \centering 

         \includegraphics[width=\linewidth]{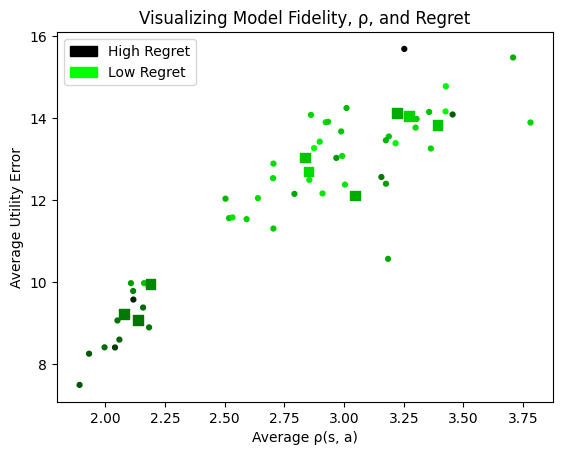} 
        
        \caption{$\rho$, $\delta(\hat{\Pi})$, and regret for \ourapproach tuning trials.
        }
        \label{fig:model_fidelity_regret}
    \end{subfigure}
    \hspace{.05cm}
     \begin{subfigure}[t]{.32\linewidth}
        \centering 

         \includegraphics[width=\linewidth]{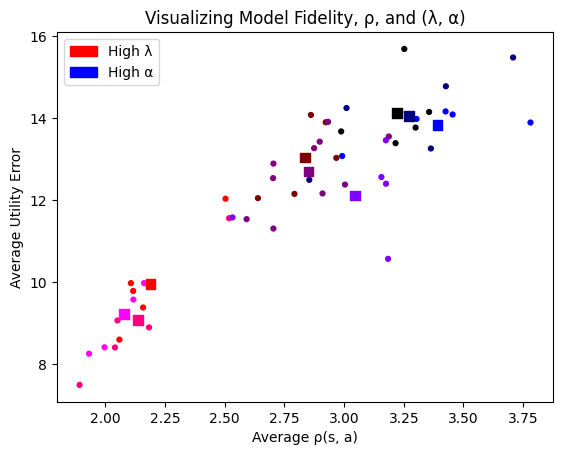}

        \caption{$\rho$, $\delta(\hat{\Pi})$, and ($\lambda$, $\alpha$) for \ourapproach tuning trials.
        }
        \label{fig:model_fidelity_parameters}
    \end{subfigure}
    \hspace{.05cm}
    \begin{subfigure}[t]{.32\linewidth}
        \includegraphics[width=\linewidth]{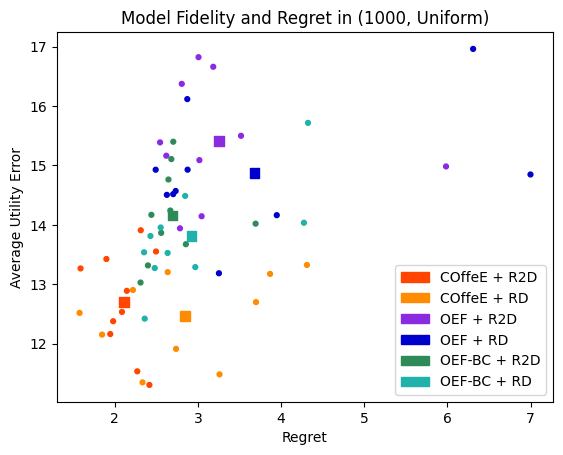}
        \caption{regret and $\delta(\hat{\Pi})$ for trials using $\mathcal{D}=(1000, \pi^{\mathcal{U}})$.}
        \label{fig:model_fidelity_approaches}
     \end{subfigure}

    \caption{Relationships among model fidelity, performance, and conservatism. 
    Each datapoint is a single trial ($\mathcal{M}_{\eval} = \text{R2D}$).
    }
    \label{fig:all_model_fidelity}
    \Description{Graphs displaying the relationship between regret, conservatism, and model fidelity.}
\end{figure*}

\textbf{Comparing with OEF: Benefits of Conservatism} - 
Except for $(1000, \sigma^{\eq})$, \ourapproach + R2D tends to outperform OEF, indicating that intentional conservatism and uncertainty quantification improve performance. 
Particularly, under a uniform policy $\pi^{\mathcal{U}}$, $\mathcal{D}$ provides relatively little information on valuable policy spaces, widening the performance margin between \ourapproach + R2D and OEF.
Under the same argument of conservatism, OEF-BC also outperforms OEF when using $\pi^{\mathcal{U}}$ datasets. 

\textbf{Comparing with OEF-BC: Strategic vs. Uninformed Conservatism} - 
\ourapproach provides consistently stronger performance than OEF-BC under all datasets except $(500, \pi^{\mathcal{U}})$.
These two approaches differ conceptually in how they apply conservatism.
Through strategy exploration, \ourapproach operates by prioritizing search in dataset-informed spaces while OEF-BC enforces similarity to dataset behavior.
These results support the notion that low-data settings provide little information on how to be conservative in a calculated manner during exploration.
Thus, \ourapproach's strategic conservatism is outperformed by OEF-BC's simple method of conservatism in low data, weak behavior policy regimes.

\textbf{Behavior Cloning in General-Sum Games} - 
We highlight that OEF-BC is substantially outperformed by all approaches under $\sigma^{\eq}$ datasets, underscoring a core property of general-sum games.
Although $\sigma^{\eq}$ provides information on several "strong" strategies, the mixture of equilibria in general-sum games is not necessarily an equilibrium itself, making mixing with a BC-trained policy arbitrarily effective.
This emphasizes a vital consideration distinguishing conservative approaches in single and multiagent environments: imitating datasets that contain strong behavior does not imply the generation of strong strategies in the latter.

\textbf{Immediate Regret Benefits of R2D} - 
We now discuss $\mathcal{M}_{\eval}$'s effect on immediately evaluated regret independent of strategy exploration.
For a given approach and dataset, R2D tends to produce lower regret solutions than RD, suggesting the optimization of conservative regret estimates benefits offline equilibrium extraction.
These findings are consistent with prior work \citep{zhang2023_offline_markov_games_general_function}.

\subsection{Analyzing Model Fidelity}

While online strategy exploration focuses on creating a set of strategies that sufficiently capture game structure, a challenge of offline game-solving is the added uncertainty of utility estimations due to limited access to data.
The extent to which our algorithm should trust its offline game model for equilibrium extraction should, intuitively, be a function of its utility approximation accuracy.
We refer to a game model's utility estimation accuracy as \term{model fidelity} and hypothesize that it is a key component for robust offline equilibrium discovery.
More precisely, since the optimization of Equation~\ref{eq:overall_response_objective} can be interpreted as skewing strategy exploration to create a higher fidelity game model, we investigate the relationship between $\rho$, parameter strength, model fidelity, and solution quality. 

For each experimental trial depicted in Fig.~\ref{fig:all_model_fidelity}, we use offline-generated strategies $\hat{\Pi}$ to reconstruct an empirical game $\hat{\Gamma}$ with true game simulations, calculating the average uncertainty $\rho$ across all simulations and average utility error across all entries in $\hat{\Gamma}$: $\delta(\hat{\Pi}) = \frac{1}{\hat{\Pi}}\sum_{\pi\in\hat{\Pi}}\sum_{i=1}^n\abs{u_i(\pi)-\hat{u}_i(\pi)}$, where $u_i(\pi)$ and $\hat{u}_i(\pi)$ are player $i$'s true game and dynamics model estimated utilities, respectively.
Figs.\ \ref{fig:model_fidelity_regret} and~\ref{fig:model_fidelity_parameters} present scatter plots relating $\rho$ and $\delta(\hat{\Pi})$, color-coded proportionally to the magnitude of regret and parameters $(\lambda, \alpha)$ applied, respectively.

Plots show a clear positive relationship between $\rho$ and $\delta(\hat{\Pi})$, indicating $\rho$ is an effective proxy for uncertainty quantification.
Parameters $\lambda$ and $\alpha$ strongly correlate with improved model fidelity but not necessarily regret.
Instead, moderate hyperparameter choices, and therefore modest model fidelity, yield the lowest regret, shown in Fig.~\ref{fig:model_fidelity_parameters}, suggesting that excessive conservatism prevents the discovery of valuable strategies even if the subgame's utility estimates are relatively accurate.
These experiments demonstrate how effective offline game-solving relies on policy generation balancing conservatism and potential strategic value (model fidelity vs. regret). 
This tradeoff between model fidelity and regret is reflected in Fig.~\ref{fig:model_fidelity_approaches}, where the use of R2D when $\mathcal{D}=(1000, \pi^{\mathcal{U}})$ tends to decrease regret across all approaches despite compromising model fidelity.

\subsection{Ablations}

\begin{figure}[h]
    \centering

    \begin{tabular}{@{}c@{}c@{}}
        \includegraphics[width=0.75\linewidth]{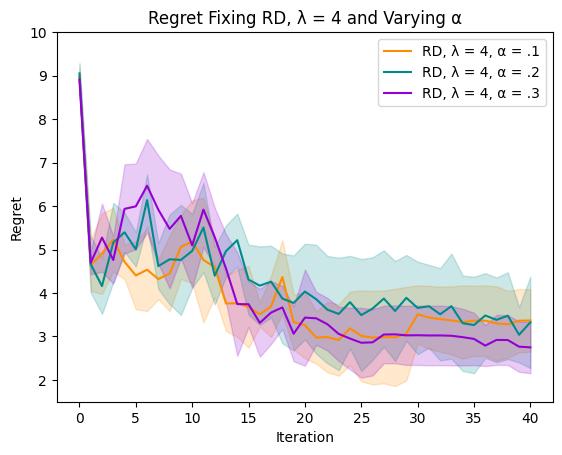} \\
        \includegraphics[width=0.75\linewidth]{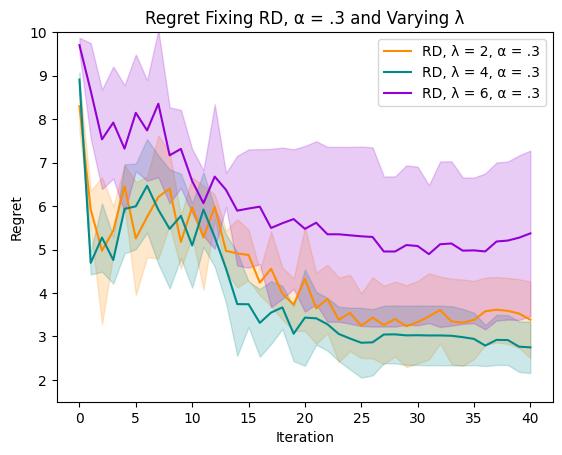} \\
        \includegraphics[width=0.75\linewidth]{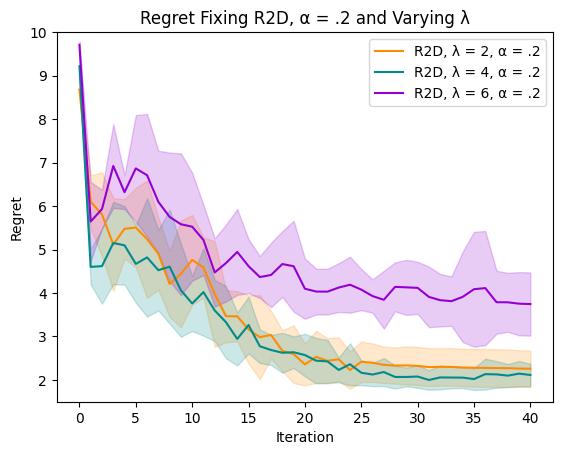} \\
        \includegraphics[width=0.75\linewidth]{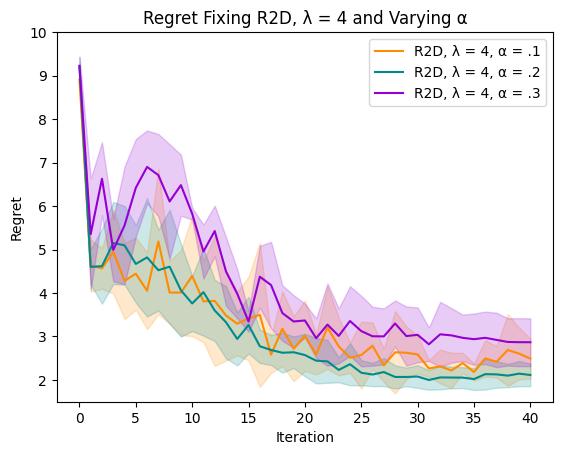} \\
    \end{tabular}


    \caption{Regret over iterations plots under various parameter settings. 
    Regret entries for $(\mathcal{M}=\text{RD}, \lambda=4, \alpha=.3)$ and $(\mathcal{M}=\text{R2D}, \lambda=4, \alpha=.2)$ are averaged over 10 trials; all others are over 5. 
    }
    \label{fig:main_paper_ablation_graph_and_table}
    \Description{Graphs displaying ablations over Coffee-PSRO hyperparameters.}
\end{figure}

We provide experiments varying $\lambda$ and $\alpha$ in Fig.~\ref{fig:main_paper_ablation_graph_and_table}, showing the subset of plots that ablate from the strongest parameter settings: ($\lambda = 4$, $\alpha=.2$) and ($\lambda=4$, $\alpha=.3$) for R2D and RD, respectively.
All remaining ablations of \ourapproach are provided in Supp.~\ref{sec:regret_tuning_results}.
Moderate levels of $\lambda$ tend to produce the strongest results, both in final regret and convergence speed. 
However, when using $\mathcal{M} =$ R2D, a more conservative MSS, smaller values of both $\lambda$ and $\alpha$ are favored.
This indicates a threshold at which excessive conservatism from all sources becomes detrimental.
Generally, higher values of $\alpha$ correlate to slower initial convergence, indicated by regret spikes in early iterations.
These spikes are commonly followed by steep dips, sometimes finding lower regret solutions than configurations with initially stronger performance.
This suggests that $\alpha$, by skewing strategy exploration toward regions with higher certainty on strategic deviations, benefits lower-regret, higher-certainty equilibrium selection in later iterations.

\begin{figure}
    \centering
    \begin{tabular}{ccccc}
        \toprule
        $\mathcal{M}$ & $\lambda \downarrow$ & $\alpha=.1$ & $\alpha=.2$ & $\alpha=.3$\\
        \midrule
        \multirow{3}{*}{RD} &  2 & $4.15 \pm 1.79$ & $3.50 \pm 0.78$ & $\textbf{3.39} \pm \textbf{0.98}$ \\
        & 4 & $\underline{3.37 \pm 0.80}$ & $\underline{3.33 \pm 1.18}$ & $\underline{\textbf{2.58} \pm \textbf{0.56}}$ \\
        & 6 & $4.91 \pm 0.61$ & $\textbf{4.03} \pm \textbf{1.35}$ & $5.37 \pm 2.13$ \\
        \midrule
        
        \multirow{3}{*}{R2D}  & 2 & $2.89 \pm 1.51$ & $\textbf{2.26} \pm \textbf{0.47}$ & $\underline{2.50 \pm 0.93}$ \\
        & 4 & $\underline{2.49 \pm 0.52}$ & $\underline{\textbf{2.12} \pm \textbf{0.27}}$ & $2.87 \pm 0.61$ \\
        & 6 & $\textbf{3.43} \pm \textbf{0.55}$ & $3.75 \pm 0.81$ & $3.64 \pm 1.01$ \\
        \bottomrule
    \end{tabular}
    \caption{Final regret under various parameter settings. 
    Regret entries for $(\mathcal{M}=\text{RD}, \lambda=4, \alpha=.3)$ and $(\mathcal{M}=\text{R2D}, \lambda=4, \alpha=.2)$ are averaged over 10 trials; all others are over 5. 
    Bold and underlined numbers indicate the lowest regret for fixed $\lambda$ and $\alpha$, respectively.
    During tuning, we use the same MSS $\mathcal{M}$ for exploration and evaluation.}
    \label{fig:placeholder}
\end{figure}

\label{sec:main_paper_ablations}

\section{Limitations}
An inherent limitation of PSRO, and by extension of \ourapproach, is the computational cost of training a sequence of deep RL strategies in sequence, which can be exorbitantly expensive for complex games.
Several approaches have focused on reducing this computational cost \citep{Bighashdel24, smith2021iterative_qmix}.
Our experiments find that \ourapproach improves over baselines if the dataset $\mathcal{D}$ provides enough information to quantify or extrapolate uncertainty in uncovered strategy spaces (strategic conservatism). 
Otherwise, \ourapproach provides no more benefit than uninformed conservatism.
Dataset coverage is especially difficult for multiagent environments, which tend to have larger search spaces than single-agent.
Lastly, experiments demonstrate the efficacy of one hyperparameter setting ($\lambda$, $\alpha$) across different datasets. 
However, it is unclear whether these same parameters generalize to other games with different reward structures and scales.

\balance 

\section{Conclusion}
We investigated a novel approach to offline game-solving, which addresses the discovery of equilibria from fixed datasets of multiagent trajectories.
These datasets may provide incomplete information on relevant strategy spaces, making game-solving strictly more difficult in the offline setting.
Under this added uncertainty, we framed this task as an equilibrium selection problem: of solutions apparently in approximate equilibrium, prefer those likely to minimize actual, true-game regret. 
We address equilibrium selection under the lens of strategy exploration, which refers to the process of deciding what strategies to include in an iteratively extended game model, by incorporating conservative algorithmic techniques from single-agent, offline RL algorithms.

Our approach, \ourapproach, extends PSRO by training a dynamics model to use in place of an online simulator.
Then, we design a modified response objective that optimizes model-quantified uncertainty given the current response target and potential strategy deviations.
Lastly, we propose a novel MSS, R2D, that optimizes conservative regret estimates.
We test our algorithm in a general-sum, sequential Bargaining game against several SOTA baselines.
Our results demonstrate that \ourapproach + R2D tends to produce lower regret solutions with higher consistency than SOTA approaches.
We also investigated the unique relationship between algorithmic components, behavior policy, and performance in general-sum games, highlighting \ourapproach's strategic method of applying conservatism.
As a final solver ($\mathcal{M}_{\eval}$), R2D tends to identify lower regret solutions than the RD baseline, underscoring the benefits of optimizing for pessimistic regret estimates in offline settings.
Lastly, experiments analyzing model fidelity and its relationship to regret demonstrate the importance of balancing conservatism and potential strategic value in strategy exploration.

\bibliographystyle{ACM-Reference-Format} 
\bibliography{sample}


\appendix

\newpage
\section{Hyperparameters}

\begin{table}[h]
    \centering
    \caption{All hyperparameters}
    \begin{tabular}{|c|c|}
        \toprule 
         Parameter Description & Chosen Parameter \\
         \midrule 
         \multicolumn{2}{|c|}{DDQN Best-Response Parameters} \\
         \midrule
         Network Width & 200 \\ 
         Network Depth & 2 \\ 
         Replay Buffer Size & 5e4 \\
         Batch Size & 64 \\ 
         Learning Rate & 1e-4 \\
         Update Target Every & 1000 \\ 
         Learn Every & 2 \\ 
         Discount Factor & .99 \\ 
         Minimum Buffer Size to Learn & 5e4 \\ 
         Epsilon Start & 1.0 \\ 
         Epsilon End & .02 \\ 
         Epsilon Linear Decay Duration & 2e5 \\
         Training Steps & 2e5 \\ 
         \midrule 
         \multicolumn{2}{|c|}{Dynamics Model Parameters} \\
         \midrule 
         Model Width &  250 \\ 
         Model Depth & 2 \\ 
         Ensemble Size & 4 \\ 
         Model Training Batch Size & 64 \\ 
         Model Training Learning Rate & 3e-4 \\ 
         Model Training Steps & 1e4 \\ 
         \midrule 
         \multicolumn{2}{|c|}{Game-Solving Parameters} \\
         Iterations & 40 \\
         Simulations per Entry & 1000 \\ 
         Steps anneal $\alpha$ & 10 \\
         \bottomrule
    \end{tabular}

    \label{tab:hyperparameters_table}
\end{table}

OEF, OEF-BC, and \ourapproach all used the same dynamics model parameters and architecture for fairness. 
Each model within an ensemble consisted of a state-transition network and a reward network, two disparate networks using the same architecture as described above.
All inputs and prediction targets were normalized with the dataset's mean and standard deviation. 
The state-transition network was trained to predict the change in state $\Delta = s_{t+1} - s_t \mid a_t$.
An observation-legal-action network was trained to share across all models within the ensemble, mapping true state to player-specific observations and legal-action masks. 
Notably, since Bargaining only consists of terminal rewards, the reward network was trained exclusively on the rewards of \textit{terminal} transitions for all players to prevent network outputs from collapsing to 0.
During training, players received a reward of 0 for all non-terminal transitions and network-predicted rewards otherwise.
Furthermore, in \ourapproach, the penalties from prediction differences $\rho$ were only applied during strategy training, not empirical game updates (where we exclusively used the ensemble's mean predictions).

We train each dynamics model using mean-squared-error loss and the parameters listed in Table~\ref{tab:hyperparameters_table}.
Observation action histories are converted to the game's defined information state through an internal module.
An episode is considered finished if either a rollout reaches a maximum length or the resulting state prediction is close (mean absolute difference < .5) to a predefined terminal state. 
Bargaining-specific details for dynamics model training are provided in Sec~\ref{sec:appendix_bargaining}.
Best-response parameters for the DDQN are provided in Table~\ref{tab:hyperparameters_table}.
These parameters were not tuned but fixed at sensible values.

\label{sec:hyperparameters}

\section{Bargaining Game Adaptation}
We designed a generalized version of the Bargaining \citep{bargaining} provided by OpenSpiel \citep{openSpiel}.
Our game is defined by several parameters, summarized in Table~\ref{tab:bargaining_parameters}.

\begin{table}[h]
    \centering
    \caption{Parameters in the bottom half are sampled at the beginning of each episode while those in the top half are held constant or define distributions that bottom-half parameters sample from.}
    \label{tab:bargaining_parameters}
    \begin{tabular}{ccp{4cm}}
        \toprule
         Symbol & Chosen & Description \\
         \midrule
         $N_{items}$ & 3 &Number of item types \\  
         $[v_{min}, v_{max}]$ & [5, 10]&Summed valuation bounds for both players $i \in \{ 1, 2\}$.\\
         $[c_{min}, c_{max}]$ & [5, 7] &Summed item count bounds for any sampled pool\\
         $T$ & 10 &Max game length \\
         $\discount$ & .99 & Discount factor of valuation \\
         \midrule
         $\{c^j\}_{j=1}^{N_{items}}$ & N/A &The number of items of each item type in the pool \\ 
         $\{v^j_i\}_{j=1}^{N_{items}}$ & N/A &Valuations for players $i \in \{ 1, 2\}$\\ 
         \bottomrule
    \end{tabular}
    
\end{table}

At the beginning of each episode, an item pool is sampled uniformly from all possible pools that satisfy two conditions:

\begin{gather*}
    c^j \geq 1 \; \forall j \in \{1 \dotsc N_{items}\} \\
    \sum_{j=1}^{N_{items}}c^j \in [c_{min}, c_{max}] 
\end{gather*}

where $c^j$ is a positive integer.
An item pool is represented by a vector of item counts $C \doteq \{c^j\}_{j=1}^{N_{items}}$.
Next, player valuations are sampled uniformly from all possible valuations that satisfy two conditions:

\begin{gather*}
    v_i^j \geq 1 \; \forall j \in \{1 \dotsc N_{items}\}, i \in \{1, 2\} \\ 
    \sum_{i=j}^{N_{items}} v_i^j \in [v_{min}, v_{max}] \; \forall i \in \{1, 2\}
\end{gather*}

where $v_i^j$ is a real number greater than or equal to 1. 
Player $i$'s valuation is represented by the vector $V_i \doteq \{ v_i^j \}_{j=1}^{N_{items}}$.
For symmetry, the first player is decided with a coin flip and valuation sample distributions are identical for both players. 
The game terminates if a player accepts an offer or the number of turns exceeds $T$.
Our chosen parameters are listed in Table~\ref{tab:bargaining_parameters}

The state, observation, information-state (history), and action vectors are represented by

\begin{gather*}
    s_t = \{\{\mathbb{I}[a_{t-1}\ = \Accept]\} + \{t\} + C + V_1 + V_2 + \{a_{t-1}\} + \{i_{curr}\} \\ 
    s_{terminal} = \{-1\}^{\mid s \in \mathcal{S} \mid}\\ 
    a_t = \{n^j\}_{j=1}^{N_{items}} \mid n^j \in \{0, \dotsc c^j\} \\ 
    o_{i, t} = \{\mathbb{I}[a_{t-1} = \Accept]\} + \{t\} + C + V_i + \{a_{t-1}\}  \\ 
    h_{i, t} = \{\mathbb{I}[a_{t-1} = \Accept]\} + \{t\} + C + V_i +\cup_{t'=0}^{t'-1}a_{t'} \\ 
\end{gather*}

where "$+$" denotes concatenation.

Note that the output of DDQN assumes a discrete action space represented by an index, not an integer vector $a_t$. 
Therefore, the network output size equals the number of all possible offers under all possible pools, computable given the parameters in Table~\ref{tab:bargaining_parameters}.
Furthermore, since the DDQN is not able to output vectorized actions as shown above, we create an internal module that converts action indices $a^{index}_t$ to offer vectors $a_t$ of length $N_{items}$.
We also create a module that converts an explicit history $\{o_{i, t'}, a_{1, t'}, a_{2, t'}\}_{t'=1}^{t}$ into a condensed information state $h_{i, t}$.

\label{sec:appendix_bargaining}

\section{Dataset Generation}
Experiments used three different dataset sizes $N \in \{500, 1000, 2000\}$ and two different behavior policies $\pi \in \{ \uniformDataset, \eqMixtureDataset \}$, referring to each dataset by the tuple $(N, \pi)$.
$\uniformDataset$ denotes a joint policy that selects legal actions at uniform random: $\uniformDataset: \mathcal{U}(\mathcal{A})$. 
$\eqMixtureDataset$ denotes a uniform strategy mixture over 5 profiles generated trials of PSRO, each consisting of 20 iterations: $\eqMixtureDataset: \mathcal{U}(\{\sigma^{\mathit{PSRO}_i}\}_{i=1}^{5})$, where uniform sampling occurs at the beginning of each episode.
For independence, all offline experiments generated independent datasets $\mathcal{D}$ at the beginning of every trial using the listed predefined joint policies.

\label{sec:dataset_generation}

\section{Other Algorithmic Details}
\label{sec:algorithm_benchmarks}

\subsection{Policy Space Response Oracles}

The PSRO algorithm \citep{lanctot2017unified_psro} iteratively extends an empirical game model $\hat{\Gamma}=(\hat{\Pi}, \hat{U}, n)$ using RL. 
We initialize $\hat{\Pi}_i$, for example, with a singleton random policy for each player $i$.
PSRO extends $\hat{\Gamma}$ at each iteration through these steps: (1)~update $\hat{U}$ by estimating payoffs for all \term{strategy profiles} (i.e., joint strategies $\hat{\Pi}$) through simulation, (2)~extract a \term{target profile} $\sigma$ by applying a \term{meta-strategy solver} (MSS) to $\hat{\Gamma}$, and (3)~derive a \term{best-response policy} $\pi^\star_i\in\Pi_i$ for each player $i$ by applying deep RL in an environment where other players are fixed to play $\sigma_{-i}$ and add $\pi^\star_i$ to $\hat{\Pi}_i$.
PSRO is summarized in Algorithm \ref{alg:psro}.

\begin{algorithm}[ht]
\caption{Policy Space Response Oracles (PSRO)}
\label{alg:psro}
\raggedright
\textbf{Input}: Meta-strategy solver $\mathcal{M}$, PSRO iterations $S$, Number of Simulations $N$\\
\textbf{Output}:  Player strategy sets $\Pi$, Player Profiles $\sigma_{S}$
\begin{algorithmic}[1] 
\STATE Initialize empty empirical game $\hat{\Gamma}: (\hat{\Pi}, \hat{U}, n)$
\STATE Initialize strategy set w/ random policy $\hat{\Pi}_i\gets\{ \pi_i^0 \}, i\in \{1,\dotsc,n\}$ 
\STATE Update empirical game $\hat{\Gamma}$ with $N$ simulations
\FOR{each PSRO iteration $s$ to $S$}
    \FOR{each player $i$ to $n$}
        \STATE Define $J_i(\theta) = \mathbb{E}_{\pi_{-i} \sim \sigma_{-i}}[\sum_{t=1}^{T} r_{i,t}]$
        \STATE $\pi^\star_i \approx  
        \arg\max_{\pi} J_i(\theta)$ 
        \STATE Update strategy sets $\hat{\Pi}_i\gets\hat{\Pi}_i \cup \pi^\star_i$
    \ENDFOR 
    \STATE Update empirical game $\hat{\Gamma}$ with $N$ simulations
    \STATE Update joint profile $\sigma_s\gets\mathcal{M}(\hat{\Gamma})$
\ENDFOR
\STATE \textbf{return} $\hat{\Pi}$, $\sigma_S$
\end{algorithmic}
\end{algorithm}

\subsection{Offline Equilibrium Finding}

\begin{algorithm}[ht]
\caption{Offline Equilibrium Finding - Behavior Cloning (OEF, OEF-BC)}
\label{alg:overall_oef_bc_algorithm}
\raggedright
\textbf{Input}: Dataset $\mathcal{D}$ of trajectories, meta-strategy solver $\mathcal{M}$, PSRO iterations $S$, Number of Simulations $N$, Behavior cloning mixing weight $\alpha_{\bc}$ \\
\textbf{Output}:  Player bc-mixed strategy sets $\Pi^{\bc}$, Player profiles $\sigma_{S}$
\begin{algorithmic}[1] 
\STATE Initialize empty empirical game $\hat{\Gamma}: (\hat{\Pi}, \hat{U}, n)$
\STATE Initialize strategy set w/ random policy $\hat{\Pi}\gets\{ \pi_i^0 \}$ 
\STATE Train ensemble dynamics model of size $K$ on $\mathcal{D}$
\STATE Train $\pi_i^{\bc}$ using behavior cloning on $\mathcal{D}$ for $i \in \{1, \dotsc n\}$ 
\STATE Update empirical game $\hat{\Gamma}$ using dynamics model with $N$ simulations.
\FOR{each PSRO iteration $s$ to $S$}
    \FOR{each player $i$ to $n$}
        \STATE Define $J_i(\theta) = \mathbb{E}_{a_t\sim(\pi_{\theta}, \sigma_{-i})}[\sum_{t=1}^{T} \overline{\mathcal{R}}_i(s_t, a_t)]$
        \STATE $\pi_i \approx \arg\max_{\pi_{\theta}} J(\theta) $ 
        \STATE Update strategy sets $\hat{\Pi}_i\gets\hat{\Pi}_i \cup \pi_{i}$
    \ENDFOR
    \STATE Update empirical game $\hat{\Gamma}$ using dynamics model with $N$ simulations
    \STATE Update joint profile $\sigma_{S}\gets\mathcal{M}(\hat{\Gamma})$
\ENDFOR

\FOR{each player $i$ to $n$}
\STATE Mix policies $\Pi^{\bc}_i = \cup_{s=1}^S \pi_i^{s,\mix}$ where $\pi_i^{s, \mix}(a \mid s) = \pi_i^s(a\mid s) (1-\alpha_{\bc}) + \alpha_{\bc} \pi^{\bc}_i(a\mid s)$ \label{alg:line:oef_bc_mix}
\ENDFOR
\STATE \textbf{return} $\Pi_i^{\bc}$, $\sigma^S$

\end{algorithmic}
\label{alg:offline_equilibrium_finding}
\end{algorithm}

Offline equilibrium finding \citep{li2022offline_equilibrium_finding}, as adapted to PSRO, is summarized in Alg.~\ref{alg:offline_equilibrium_finding}. 
The approach trains a dynamics model on dataset $\mathcal{D}$ and uses it in place of an online simulator. 
In our approach, we use the ensemble architecture and Double Deep Q-Learning as described in Sections \ref{sec:train_dynamics_model} and \ref{sec:implementation_and_reproducibility}, respectively, for a fair comparison with \ourapproach.
OEF also trains a behavior cloned policy $\pi^{\bc}$ on $\mathcal{D}$, which is used at the end of training to mix with the output policy $\pi \in \Pi$ as indicated by Line \ref{alg:line:oef_bc_mix}.
Note that mixing implies sampling action $a_t$ from $\pi^{\bc}_i$ or $\pi^s_i$ at each timestep.
\citet{li2022offline_equilibrium_finding} chooses $\alpha_{\bc}$ \textit{at test time} using a true game simulator to determine which setting has the lowest exploitability.
While this can be seen as a violation of the offline learning assumption, we evaluate \ourapproach against OEF-BC($\alpha_{\bc}$) with the strongest test-time results. 
The OEF baseline is identical to OEF-BC but sets $\alpha_{\bc}$ to 0.

\section{True Game Regret}
Offline-generated profiles are evaluated in the true game. 
We train an online best response to profiles outputted at the last $S_{\eval}$ iterations: $\Pi_i^{online} = \{ \BR_i(\sigma_{i, s}) \mid s \in \{T - S_{\eval}, \dotsc T \}\}$, where $s$ indicates the $s$-th profile generated and $T$ denotes the number of PSRO iterations.
We construct a combined empirical game that includes offline and online trained strategies $\Pi^{\eval}_i = \Pi_i \cup \Pi_i^{online}$.
Then we construct a normal-form game using $\Pi_{\eval}$, recomputing all entries using $N=1000$ \textit{true game} simulations.
Player $i$'s regret given iteration $s$'s joint profile is approximated by the following:

\begin{displaymath}
    \Regret_i(\sigma_s) \approx \max_{\pi_{i}^{\eval} \in \Pi_i^{\eval}} u_i(\pi_i^{\eval}, \sigma_{-i,s}) - u_i(\sigma_s)
\end{displaymath}

All results report the sum of regret across players $\sum_{i=1}^n \Regret_i(\sigma_s)$.
Ablations use $S_{\eval} = T = 40$ while all other experiments use $S_{\eval} = \frac{T}{2} = 20$.

\label{sec:true_game_regret_calculations}
\section{Additional Results}
\label{sec:regret_tuning_results}

\subsection{Regret Results During Tuning}

\begin{table}[h]
    \centering
    \caption{True game regret values of final profiles outputted by baseline runs. Each trial consists of 40 iterations, where an iteration is one new strategy, and reported values are averaged over 10 trials. Note that when $\alpha_{\bc} = 0.0$, we refer to the baseline named OEF.}
    \begin{tabular}{ccc}
    \toprule
    $\alpha_{\bc}$ & OEF-BC + RD & OEF-BC + R2D\\ 
    0.0 & $3.74 \pm 2.10$ & $3.25 \pm 1.00$ \\ 
    0.2 & $2.96 \pm 1.07$ & $2.70 \pm 0.39$\\ 
    0.4 & $3.49 \pm 0.25$ & $3.50 \pm 0.22$ \\ 
    0.6 & $4.79 \pm 0.25$ & $4.87 \pm 0.19$ \\ 
    0.8 & $6.56 \pm 0.28$ & $6.70 \pm 0.19$ \\ 
    \bottomrule
    \end{tabular}
    \label{tab:ablation_final_regret_table}
\end{table}

\begin{figure*}[ht]
    \centering
    
    \begin{tabular}{@{}c@{\hspace{.1cm}}c@{\hspace{.1cm}}c@{}}
        \includegraphics[width=0.32\linewidth]{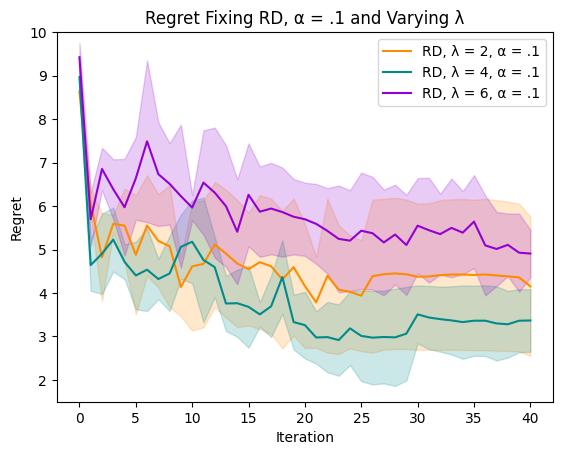} &  
        \includegraphics[width=0.32\linewidth]{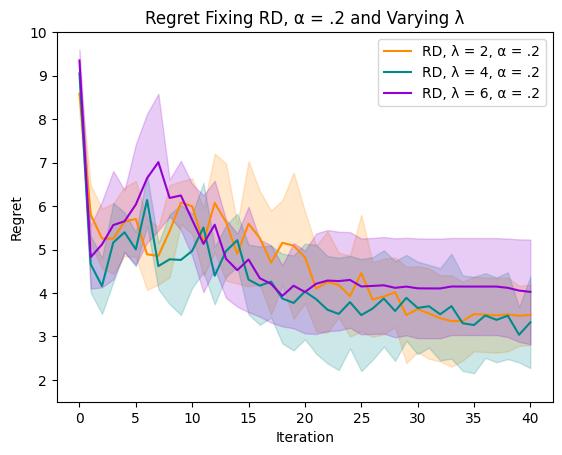} &
        \includegraphics[width=0.32\linewidth]{Images/Ablations/Replicator_Dynamics/Regret_Fixing_RD,_alpha_=_.3_and_Varying_lambda.jpg} \\
        
        \includegraphics[width=0.32\linewidth]{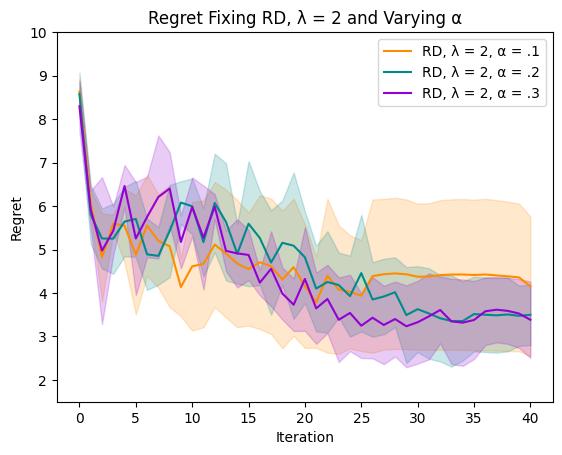} &  
        \includegraphics[width=0.32\linewidth]{Images/Ablations/Replicator_Dynamics/Regret_Fixing_RD,_lambda_=_4_and_Varying_alpha.jpg}&
        \includegraphics[width=0.32\linewidth]{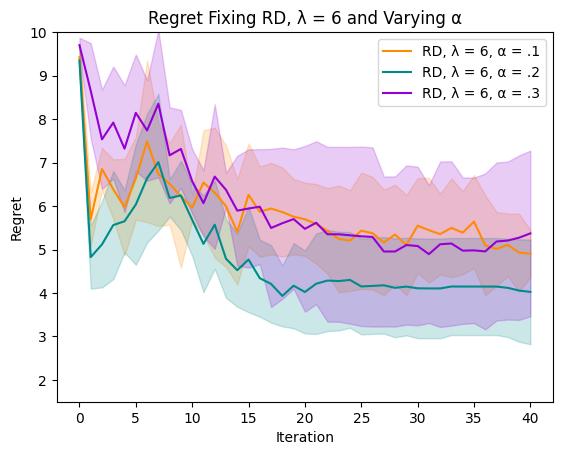}\\

        \includegraphics[width=0.32\linewidth]{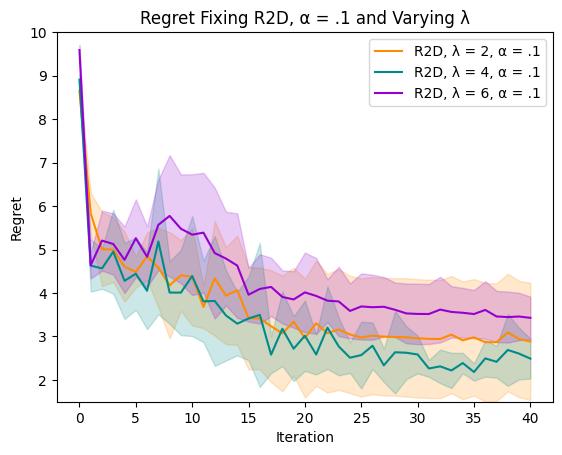} &  
        \includegraphics[width=0.32\linewidth]{Images/Ablations/Robust_Replicator_Dynamics/Regret_Fixing_R2D,_alpha_=_.2_and_Varying_lambda.jpg} &
        \includegraphics[width=0.32\linewidth]{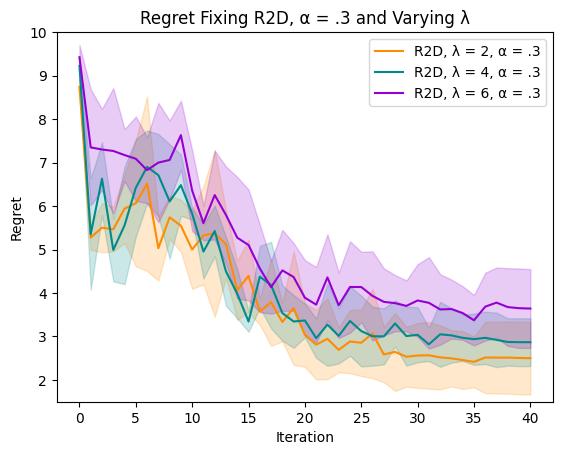} \\
        
        \includegraphics[width=0.32\linewidth]{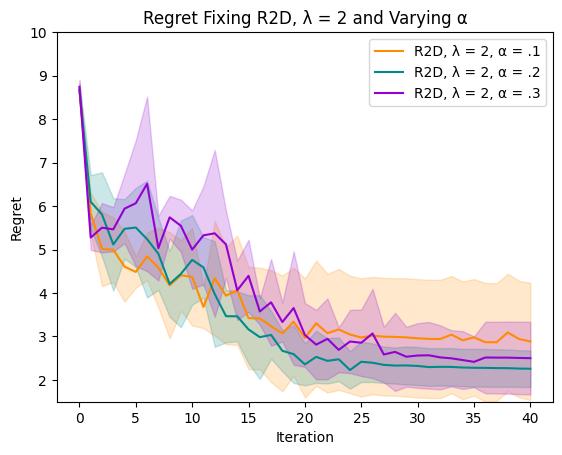} &  
        \includegraphics[width=0.32\linewidth]{Images/Ablations/Robust_Replicator_Dynamics/Regret_Fixing_R2D,_lambda_=_4_and_Varying_alpha.jpg}&
        \includegraphics[width=0.32\linewidth]{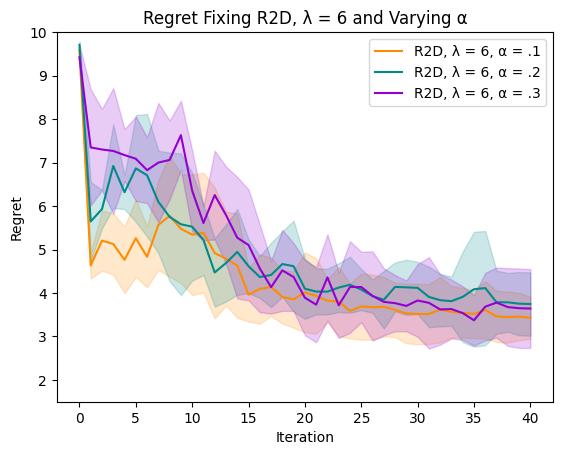}\\

    \end{tabular}
    \caption{Ablation over \ourapproach parameters. The top two rows ablate over $\alpha$ and $\lambda$, fixing $\mathcal{M}$ to RD, while the next two rows fix $\mathcal{M}$ to R2D.}
    \Description{Graphs displaying additional ablations over Coffee-PSRO hyperaparameters.}
    \label{fig:ablation_parameters}
\end{figure*}

We provide final true game regret values for tested baselines in Table~\ref{tab:ablation_final_regret_table} and all ablations over \ourapproach parameters ($\lambda$, $\alpha$, $\mathcal{M}$) in Table~\ref{tab:appendix_tuning_coffee} and Fig.~\ref{fig:ablation_parameters}.

\begin{table}[h]
    \centering
    \caption{True game regret values of final profiles outputted by \ourapproach during tuning. Underlined values are averaged over 10 trials, 5 otherwise.}
    \begin{tabular}{c@{\hspace{.2cm}}c@{\hspace{.2cm}}c@{\hspace{.3cm}}c@{\hspace{.3cm}}c@{\hspace{.3cm}}}
        \toprule
        $\mathcal{M}$ & $\lambda \downarrow$ & $\alpha = .1$ & $\alpha = .2$ & $\alpha = .3$ \\
        \midrule
        \multirow{3}{*}{RD} 
            & 2 & $4.15 \pm 1.79$ & $3.50 \pm 0.77$ & $3.39 \pm 0.98$ \\
            & 4 & $3.37 \pm 0.80$ & $3.33 \pm 1.18$ & $\underline{2.64 \pm 0.55}$ \\
            & 6 & $4.91 \pm 0.61$ & $4.03 \pm 1.35$ & $5.37 \pm 2.13$ \\
        \midrule
        \multirow{3}{*}{R2D} 
            & 2 & $2.89 \pm 1.51$ & $2.26 \pm 0.47$ & $2.50 \pm 0.93$ \\
            & 4 & $2.49 \pm 0.52$ & $\underline{2.12 \pm 0.27}$ & $2.87 \pm 0.61$ \\
            & 6 & $3.43 \pm 0.55$ & $3.75 \pm 0.81$ & $3.64 \pm 1.01$ \\
        \bottomrule
    \end{tabular}
    \label{tab:appendix_tuning_coffee}
\end{table}

\end{document}